\documentclass[11pt,a4paper]{article}

\usepackage[T1]{fontenc}
\usepackage[english]{babel}

\usepackage[margin=1in]{geometry}

\usepackage{amsmath,amssymb}

\usepackage{multirow}
\usepackage{booktabs}
\usepackage{threeparttable}
\usepackage{makecell}
\usepackage{subcaption}

\usepackage{graphicx}
\usepackage{xcolor}
\usepackage{tabularx}
\usepackage{longtable}
\usepackage{dsfont}
\usepackage{placeins}
\usepackage{tikz}
\usetikzlibrary{positioning,fit,backgrounds,arrows.meta,calc,shapes.geometric,decorations.pathreplacing,matrix}

\usepackage[hidelinks]{hyperref}

\usepackage[numbers,square,sort&compress]{natbib}
\hypersetup{
  pdftitle={What Do Mayan Embeddings Capture?},
  pdfauthor={Silvia Fernandez-Sabido and Jorge Reyes-Magana},
  pdfkeywords={Low-resource NLP, Yucatec Maya, Graph-based Embeddings, Semantic Relational Structures}
}

\begin{document}

\title{What Do Mayan Embeddings Capture? Overcoming the Long-Tail through Graph Topologies and Native Expert Validation}

\author{
  Silvia Fernández-Sabido\thanks{Email: \href{mailto:sfernandez@centrogeo.edu.mx}{sfernandez@centrogeo.edu.mx}.} \\
  \small Secretariat of Science, Humanities, Technology, and Innovation (SECIHTI), commissioned to the\\
  \small Center for Research in Geospatial Information Sciences (CentroGeo),\\
  \small Carretera Sierra Papacal, Chuburná Puerto, Sierra Papacal, 97302 Mérida, Yucatán, Mexico
  \and
  Jorge Reyes-Magaña\thanks{Corresponding author. Email: \href{mailto:jorge.reyes@correo.uady.mx}{jorge.reyes@correo.uady.mx}.} \\
  \small Faculty of Mathematics, Autonomous University of Yucatán,\\
  \small Anillo Periférico Norte, Tablaje Cat. 13615, 97205 Mérida, Yucatán, Mexico
}

\date{\today}
\maketitle

\begin{abstract}
\noindent
Word embeddings are essential for modern NLP, yet their development for low-resource languages is hindered by data sparsity and complex morphology. This study addresses the question: What do Mayan embeddings capture? We present the first human-rated gold standard of semantic representations for Yucatec Maya, a morphologically rich language, by introducing a graph-based methodology designed to overcome the long-tail distribution of rare inflected forms. Leveraging a multi-source corpus of 717,471 tokens, we propose an approach using Node2Vec to capture global topological relations, comparing it against traditional local-window baselines. Evaluation was conducted through a rigorous protocol involving five native linguistic experts and a curated gold standard of 90 word pairs. Our results demonstrate that graph-based embeddings significantly outperform distributional methods, achieving a robust Spearman correlation of $\rho = 0.593$ ($p < 10^{-9}$) on the full gold standard, which rises to $\rho = 0.807$ ($p < 10^{-4}$) on high-consensus pairs. 
In contrast, Word2Vec yields non-significant results  ($\rho = 0.170, p > 0.10$), while FastText exhibits non-significant, negative correlations ($\rho = -0.080, p > 0.45$), suggesting that character n-grams introduce morphological noise in small-scale Mayan corpora. Qualitative analysis confirms that graph topologies effectively bridge relational gaps where co-occurrence data is sparse. We release our embeddings, the first expert-validated gold standard, and the processed corpus to support the digital revitalization of the Mayan language family.

\end{abstract}

\section{Introduction}\label{sec1}

The development of Natural Language Processing (NLP) technologies has disproportionately favored high-resource languages, leaving more than 90\% of the world's linguistic diversity in a state of digital under-representation \citep{joshi2020state}. This disparity is particularly acute for the Indigenous languages of the Americas, which face a combination of data sparsity, complex morphology, and orthographic fragmentation \citep{bird2020decolonising,mager2018challenges}. Yucatec Maya, spoken by approximately 775,000 people in Mexico \citep{inegi2020}, is a vital language with a growing digital presence, yet it lacks the validated semantic resources necessary for modern computational applications.

Traditional word embedding architectures rely on massive distributional evidence to converge. In low-resource settings, these models often suffer from ``dimensional collapse'', where the lack of context leads to unstable or semantically hollow representations. This work addresses this challenge by proposing a graph-based methodology for semantic modeling. By aggregating local co-occurrences into a global topological structure, we capture multi-order semantic relations that remain invisible to window-based models.

In light of these challenges, we address the following research question: \textit{What do Mayan embeddings capture when global graph topologies are used to bridge the long-tail of morphological complexity?}

To answer this, our main contributions are: (1) the construction of a curated 717,471-word multi-source corpus for Yucatec Maya; (2) the development of a human-in-the-loop validation protocol involving native linguistic experts; and (3) the release of the first validated semantic gold standard (90 pairs) and high-quality embeddings for the language. Our results demonstrate that graph-based models provide a superior and more stable semantic signal, particularly on high-consensus pairs where traditional baselines fail to achieve significance.

\section{Related Work} \label{sec:relwork}

The landscape of semantic representations has evolved from count-based vectors to dense distributed embeddings like GloVe \citep{pennington2014glove} and Word2Vec \citep{mikolov2013efficient,mikolov2013distributed}. While these methods are robust for languages with billions of tokens, their application to low-resource and morphologically rich languages remains a significant challenge.

\subsection{NLP for Indigenous Languages} 
Research on the languages of the Americas has highlighted the critical need for decolonizing speech and language technologies \citep{bird2020decolonising}. Recent efforts, such as the AmericasNLP shared tasks \citep{mager2021findings, chiruzzo2024findings}, have focused on challenges like machine translation for polysynthetic languages. However, the intrinsic evaluation of semantic spaces for Yucatec Maya has been largely overlooked due to the absence of expert-validated benchmarks.

\subsection{Morphological Complexity} 
Yucatec Maya is characterized by rich morphology. For such languages, traditional models often fail to generalize across inflected forms. While sub-word information in FastText \citep{bojanowski2017enriching} or neural morphological segmentation \citep{kann2017oneshot} attempt to address this, they still require substantial data to avoid capturing noise.

\subsection{Graph-Based Embeddings}
To overcome data sparsity, recent studies have transitioned from linear windows to global topology using methods like DeepWalk \cite{perozzi2014deepwalk}, LINE \cite{tang2015line}, and Node2Vec \cite{grover2016node2vec}. These methods treat the corpus as a network, allowing the model to explore latent semantic paths through random walks. 
\citeauthor{gutierrezvasques2017lowresourcebilinguallexiconextraction} applied Node2Vec to a bilingual graph built from a small Spanish–Nahuatl parallel corpus and showed that the resulting graph-based embeddings considerably outperformed Word2Vec for bilingual lexicon extraction under low-resource conditions.

\subsection{Semantic Evaluation} 
The evaluation of word embeddings typically relies on similarity tasks like SimLex-999 \citep{hill2015simlex}. However, \citet{faruqui2016problems} noted significant problems with these tasks in representing true semantic nuances. 

In the context of American Indigenous languages, \citet{gongora-etal-2022-use} evaluated Guaraní embeddings using a translated MC-30 dataset. Similarly, recent attempts to model Mayan semantics \citep{molina2025generating} demonstrated that while Swadesh lists provide stable correlations, the MC-30 dataset remains significantly more challenging for distributional models. 

A critical limitation of these approaches is their dependence on linear translations of datasets like Swadesh and MC-30, which are fundamentally rooted in Anglocentric semantic frameworks. Often adapted into Spanish before being translated into Indigenous languages, these benchmarks frequently impose foreign conceptual mappings that fail to reflect the unique relational nuances of the Mayan worldview. 

To address these limitations, our study leverages the relational structure of a global co-occurrence graph to identify word pairs representing a wide spectrum of semantic proximity. Rather than relying on translated benchmarks, these pairs were scored by native linguistic experts to ensure cultural and contextual accuracy. We then employ the Spearman rank correlation to evaluate model performance against these expert-tailored scores, ensuring that the evaluation reflects the linguistic reality of Yucatec Maya.

\section{Corpus Construction} \label{sec:corpus}

\subsection{Multi-Source Composition}

To mitigate the bias inherent in single-source datasets, we compiled a multi-genre corpus of Yucatec Maya comprising 843,019 tokens across 116,824 sentences. As detailed in Table \ref{tab:corpus_stats}, the collection spans five distinct domains: journalistic (La Jornada Maya), conversational (T'aantsil), general interest (Sedeculta), scientific dissemination (Ecofronteras), and literary texts (UNAM).

The distribution is primarily driven by journalistic content, which accounts for 60.6\% of the total word count. However, the inclusion of scientific and literary sources ensures a higher lexical diversity, capturing formal and technical registers that are often absent in social media-mined corpora. 

\begin{table}[t]
\centering
\caption{Quantitative overview of the Yucatec Maya corpus.}
\label{tab:corpus_stats}

\small
\begin{tabular}{llrr}
\toprule
Source & Genre & Sents & Words \\
\midrule
Jornada Maya\footnotemark[1] & Journalistic & 58,935 & 511,513 \\
T'aantsil\footnotemark[2] & Conversational & 31,790 & 166,215 \\
Sedeculta\footnotemark[3] & General & 14,410 & 75,732 \\
Ecofronteras\footnotemark[4] & Scientific & 7,173 & 60,309 \\
UNAM\footnotemark[5] & Literary & 4,516 & 29,250 \\
\midrule
Total & --- & 116,824 & 843,019 \\
\bottomrule
\end{tabular}
\footnotesize
\footnotetext[1]{https://www.lajornadamaya.mx/k'iintsil}
\footnotetext[2]{https://taantsil.com.mx/}
\footnotetext[3]{Used in AmericasNLP 2024 Shared Task \citep{chiruzzo2024findings}}
\footnotetext[4]{ https://revistas.ecosur.mx/ecofronteras/index.php/eco}
\footnotetext[5]{http://www.corpus.unam.mx/}
\end{table}

\subsection{Preprocessing and Linguistic Normalization} \label{sec:normalize}

Digital Yucatec Maya faces significant orthographic fragmentation. 
All sources underwent a rigorous normalization pipeline:

\textbf{Orthographic Standardization:} We applied NFC Unicode normalization and standardized glottal stop variants to the standard apostrophe (U+0027). This prevents a single morpheme from being fragmented into multiple entities due to encoding noise.

\textbf{Spanish Interference Control:} We implemented a sentence-level filter, removing entries where Spanish content exceeded 60\%. This purged dominant Spanish segments while preserving naturalized loanwords and authentic code-switching.

\textbf{Structural Cleaning:} Sentences with fewer than five words were discarded to ensure sufficient context for co-occurrence extraction.

\subsection{Resulting Corpus Statistics and Sparsity}

After refinement, the final corpus consists of 717,471 tokens. A key challenge for representational learning is the Type-Token Ratio (TTR) of 0.085. With 60,957 unique types and an average frequency of only 12 appearances per type, the dataset exhibits extreme lexical sparsity.

This long-tail distribution confirms that Yucatec Maya’s morphological complexity triggers a high rate of \textit{hapax legomena}, consistent with Zipfian distributions in polysynthetic and highly productive languages \citep{bentz2016comparison,czarnowska-etal-2019-dont}. Such sparsity poses a fundamental challenge for local-window models like Word2Vec, which require multiple observations to estimate stable vectors. This characteristic justifies our transition toward a graph-based approach (Node2Vec), capable of leveraging global topological connections to bridge these relational gaps and capture semantic dependencies that would otherwise remain unreachable in sparse data settings.

\section{Methodology}

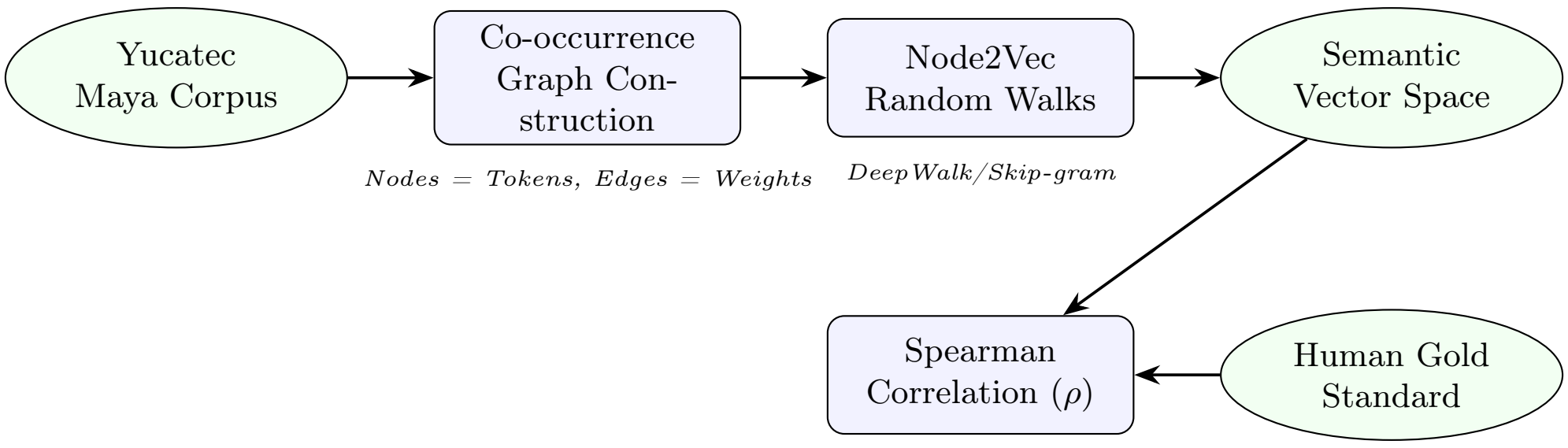
\begin{figure}[t]
\centering
\begin{tikzpicture}[
    scale=0.90,
    transform shape,
    node distance=1.5cm and 0.8cm, 
    block/.style={rectangle, draw, fill=blue!5, text width=2.6cm, text centered, rounded corners, minimum height=1.1cm, font=\footnotesize},
    data/.style={ellipse, draw, fill=green!5, text width=2.0cm, text centered, minimum height=0.9cm, font=\footnotesize},
    arrow/.style={-{Stealth[scale=1.0]}, line width=0.7pt}
]
\node (corpus) [data] {Yucatec Maya Corpus};
\node (graph) [block, right=of corpus] {Co-occurrence Graph Construction};
\node (n2v) [block, right=of graph] {Node2Vec Random Walks};
\node (vectors) [data, right=of n2v] {Semantic Vector Space};
\node (human) [data, below=of vectors] {Human Gold Standard};
\node (eval) [block, left=of human] {Spearman Correlation ($\rho$)};
\draw [arrow] (corpus) -- (graph);
\draw [arrow] (graph) -- (n2v);
\draw [arrow] (n2v) -- (vectors);
\draw [arrow] (vectors) -- (eval);
\draw [arrow] (human) -- (eval);
\node[below=0.1cm of graph, font=\fontsize{6}{8}\selectfont\itshape] {Nodes = Tokens, Edges = Weights};
\node[below=0.1cm of n2v, font=\fontsize{6}{8}\selectfont\itshape] {DeepWalk/Skip-gram};
\end{tikzpicture}
\caption{Architecture of the proposed semantic evaluation pipeline. Rectangular blue nodes represent computational processes, while green elliptical nodes denote data objects.}
\label{fig:all_process}
\end{figure}

The proposed architecture, illustrated in Figure \ref{fig:all_process}, transitions from corpus data to a structured semantic vector space through a graph-based pipeline.

\subsection{Co-occurrence Graph Construction}

While both Word2Vec and our approach utilize a sliding window, they differ fundamentally in how they process co-occurrence data. Unlike local-window distributional models that treat each window as an independent training instance, we aggregate the entire corpus topology into a global weighted undirected graph $G = (V,E)$.
Nodes $V$ represent unique word types identified after the normalization process described in Section \ref{sec:normalize}. 

%
%
The graph is populated by iterating through each sentence
$S = (s_1, \dots, s_{|S|})$ in the corpus $\mathcal{C}$, where $s_k$
denotes the word type at position $k$. The weight of the edge between
two distinct word types $i$ and $j$ is the number of times they
co-occur within a window of size $L$:
\begin{equation}
  w_{ij} = \sum_{S \in \mathcal{C}} \;
           \sum_{\substack{1 \le k, m \le |S| \\ k \neq m}}
           \mathbb{1}\!\left( s_k = i \,\wedge\, s_m = j \,\wedge\, |k - m| \le L \right),
  \label{eq:cooc}
\end{equation}
where $\mathbb{1}(\cdot)$ is the indicator function, which equals 1 if
the condition holds and 0 otherwise. Since $w_{ij} = w_{ji}$, the
resulting graph is undirected, and an edge $(i, j) \in E$ exists
whenever $w_{ij} > 0$.

By using a window size of $L=10$, we ensure that the graph captures multi-order semantic relations\footnote{The choice of $L=10$ follows standard practices for capturing broader semantic and thematic relations rather than purely syntactic ones \cite{jurafsky2026speech}. Given the highly agglutinative nature of Yucatec Maya and its flexible word order, a wider window ensures that long-distance dependencies are not omitted during the graph construction.}. Unlike Word2Vec, which requires high frequency to estimate stable local vectors, our graph-based approach leverages the global connectivity of the network, allowing even low-frequency long-tail terms to derive their meaning from their position within the entire topological structure.

The final network is exported in GEXF (Graph Exchange XML Format) with UTF-8 encoding to preserve the specificities of Maya orthography, such as glottal stops and saltillos. The resulting topology reflects the high connectivity and sparsity of the language, with properties summarized in Table \ref{tab:graph_properties}. Despite a low graph density ($1.14 \times 10^{-3}$), the average degree of $\sim$69.8 indicates that the graph successfully bridges the gap between sparse lexical entries through a well-connected core of co-occurrences.
\begin{table}[ht]
\caption{Properties of the resulting co-occurrence graph.}
\begin{tabular}{@{}l r@{}} 
\toprule
\textbf{Property} & \textbf{Value} \\
\midrule
Nodes ($|V|$) & 60,957 \\
Edges ($|E|$) & 2,130,145 \\
Average degree & $\sim$69.8 \\
Graph density & $1.14 \times 10^{-3}$ \\
Encoding/Format & UTF-8 / GEXF \\
\bottomrule
\end{tabular}

\label{tab:graph_properties}
\end{table}
\subsection{Node2Vec Embeddings}

To learn semantic representations, we employ Node2Vec, a graph embedding technique that extends the Word2Vec skip-gram architecture to networks through biased random walks. 
This method is particularly suited for our corpus for two reasons: 
1) Multi-order Connectivity: Through random walks, the model captures ``friend-of-friend'' relationships, discovering semantic similarities between words that may not co-occur directly but share similar topological neighborhoods.
2) Structural Stability: By exploring the global graph rather than local sentence windows, Node2Vec mitigates the impact of low-frequency inflected forms that typically cause ``dimensional collapse'' in Word2Vec and FastText models.
The final vector space is generated by training a Skip-gram model on the sequences generated by these walks, resulting in a dense representation that we subsequently evaluate against the human gold standard. The embeddings trained in this study are publicly available on GitHub\footnote{https://github.com/Yakunaj/graph-embeddings}, along with the evaluation data provided by expert linguists.

\section{Evaluation Framework}

The validation of semantic representations in low-resource contexts requires a transition from purely computational metrics to expert-driven frameworks. 
As illustrated in Figure \ref{fig:human_validation}, we implemented a human-in-the-loop pipeline to curate a gold standard for Yucatec Maya.

\textbf{Candidate Generation and Stratification:} We extracted the 300 nearest neighbors for 50 seed words using a $300d$ Node2Vec model. 
This dimensionality was selected based on the stability analysis in Table \ref{tab:stability_results}, which shows the vector space reaching a ``High'' stability plateau ($J \approx 0.57$) starting at $150d$. This ensures that the semantic neighbors used for gold standard construction are structural properties of the graph rather than stochastic artifacts.
These seed words were pre-selected by linguists to cover core social, natural, and grammatical domains. 
To ensure the evaluation captured a broad spectrum of semantic relations, pairs were stratified into three cosine similarity tiers: \textit{High} ($>0.7$), \textit{Medium} ($0.4$--$0.7$), and \textit{Low} ($<0.4$).

\textbf{Expert Filtering and Noise Reduction:} A critical phase involved a manual review by a native-speaking linguist to purge corpus-induced artifacts. Out of $109$ initial candidates, $19$ pairs ($17\%$) were excluded. These exclusions (summarized in Table \ref{tab:exclusion_categories}) primarily targeted non-naturalized Spanish loanwords, orthographic noise from diverse media sources, and proper names that do not reflect productive semantic clusters.

\textbf{Final Validation Protocol:} The resulting $90$ curated pairs were evaluated by five native linguists under a blind protocol. Using a $0$--$4$ Likert scale, experts rated the semantic strength without access to model scores, where $0$ indicates no semantic relation and $4$ represents perfect synonymy or identity. 
The final human-assigned scores were then correlated with model predictions using the Spearman rank correlation coefficient ($\rho$). Statistical significance is assessed at the $\alpha = 0.05$ level. However, to account for the robustness of the graph-based representations, we also report and highlight results that meet the more stringent $\alpha = 0.01$ threshold.
This process yielded a high-quality gold standard with an aggregate mean score and consensus measurement, serving as the ground truth for our correlation analyses in Section \ref{sec:results}.

\begin{figure}[t]
\centering
\begin{tikzpicture}[
    node distance=1.0cm and 0.8cm, 
    base/.style={draw, thick, align=center, font=\footnotesize\sffamily, minimum height=0.8cm, inner sep=4pt},
    input/.style={base, fill=gray!10, rounded corners=2pt},
    proc/.style={base, fill=blue!5, minimum width=3cm},
    human/.style={base, fill=orange!5, diamond, aspect=2, inner sep=0pt, minimum width=3.2cm},
    output/.style={base, fill=green!5, thick, double, rounded corners=2pt},
    arrow/.style={-{Stealth[scale=1.2]}, thick}
]
\node[input] (emb) {Node2Vec Embeddings \\ ($d=300$)};
\node[input, right=of emb] (seeds) {50 Expert Seeds \\ (3 Categories)};
\node[proc, below=1.2cm of $(emb.south)!0.5!(seeds.south)$] (knn) {300-NN Extraction \& \\ Stopword Filtering};
\node[proc, below=of knn] (strat) {\textbf{Stratification} \\ \scriptsize High ($>0.7$) $|$ Med ($0.4-0.7$) $|$ Low ($<0.4$)};
\node[human, below=of strat] (expert) {Expert Linguistic \\ Review};
\node[input, right=1.2cm of expert, text width=3cm] (noise) {Excluded (19 pairs) \\ \scriptsize (Loanwords, Orthography, \\ Proper Names)};
\node[proc, below=of expert] (cons) {Multi-Expert Blind Rating \\ (5 Native Speakers)};
\node[output, below=of cons] (gold) {\textbf{Final Semantic Gold Standard} \\ (90 Curated Pairs)};
\draw[arrow] (emb.south) -- ++(0,-0.4) -| (knn.north);
\draw[arrow] (seeds.south) -- ++(0,-0.4) -| (knn.north);
\draw[arrow] (knn) -- (strat);
\draw[arrow] (strat) -- (expert);
\draw[arrow] (expert) -- (cons) node[midway, right, font=\scriptsize] {109 $\to$ 90};
\draw[arrow, dashed] (expert) -- (noise);
\draw[arrow] (cons) -- (gold);
\end{tikzpicture}
\caption{The Human-in-the-loop Validation Pipeline. This process refines computational embeddings into a curated gold standard by stratifying similarity levels and filtering corpus artifacts through expert linguistic intervention.}
\label{fig:human_validation}
\end{figure}
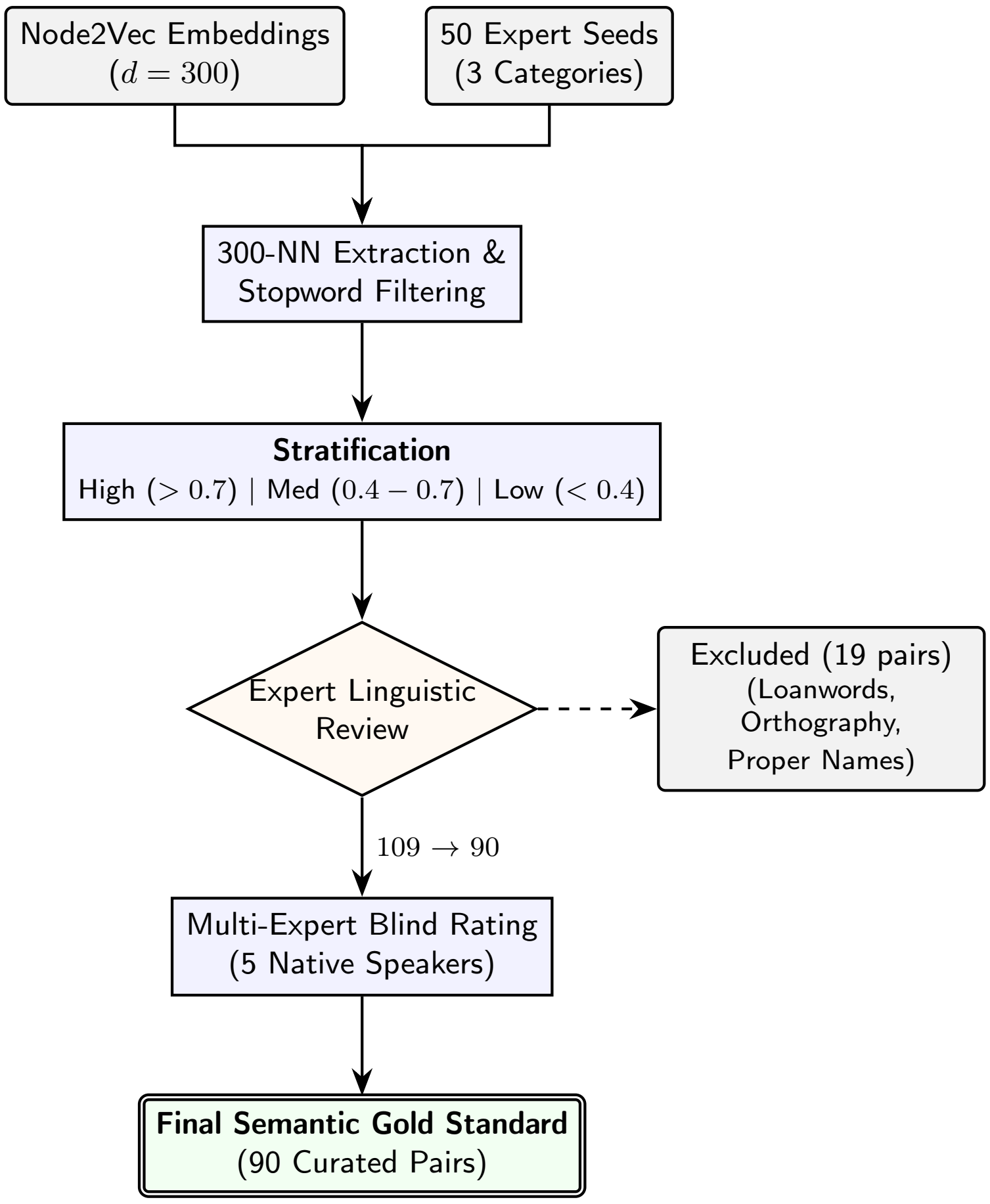
\begin{table}[h]
\centering
\caption{Jaccard Stability across Dimensions}
\begin{tabular}{lc p{2.5cm}}
\toprule
\textbf{Comparison ($d_1, d_2$)} & \textbf{Avg. $J$} & \textbf{Stability} \\
\midrule
$25d$ vs $50d$   & $0.282$ & Low \\
$50d$ vs $100d$  & $0.396$ & Moderate \\
$\textbf{100d}$ \textbf{vs} $\textbf{150d}$  & $\textbf{0.543}$ & \textbf{Acceptable} \\
$150d$ vs $200d$ & $0.582$ & High \\
$200d$ vs $250d$ & $0.572$ & High \\
$250d$ vs $300d$ & $0.567$ & High \\
\bottomrule
\end{tabular}

\label{tab:stability_results}
\end{table}
%
%
%
\begin{table}[ht]
\centering
\caption{Summary of 19 excluded word pairs.}
\begin{tabular}{lc p{2.7cm}} 
\toprule
\textbf{Category} & \textbf{Qty} & \textbf{Examples} \\
\midrule
Foreign/Proper names & 4 & \textit{orkun, hamuriye} \\
Spelling errors      & 6 & \textit{jahc, yeetel} \\
Loanwords/Acronyms   & 3 & \textit{batallón, pecda} \\
Residual functional & 6 & \textit{ti'al} \\
\midrule
\textbf{Total}       & \textbf{19} & \\
\bottomrule
\end{tabular}
\label{tab:exclusion_categories}
\end{table}

\section{Results and Analysis} \label{sec:results}

The evaluation demonstrates a clear superiority of graph-based embeddings over traditional distributional models in low-resource settings. As shown in Table \ref{tab:main_results}, Node2Vec achieves a robust correlation with expert judgment ($\rho = 0.593, p < 10^{-9}$), while Word2Vec and FastText fail to maintain statistical significance as dimensionality increases.
\begin{table}[t]
\centering
\footnotesize
\caption{Main Results: Comparative performance across dimensions and consensus levels. Node2Vec consistently outperforms distributional baselines, especially when filtering for expert agreement.}
\begin{tabular*}{\textwidth}{@{\extracolsep{\fill}} l r cc cc cc @{}}
\toprule
 & & \multicolumn{2}{c}{\textbf{Node2Vec (Ours)}} & \multicolumn{2}{c}{\textbf{Word2Vec (SG)}} & \multicolumn{2}{c}{\textbf{FastText}} \\
\cmidrule(lr){3-4} \cmidrule(lr){5-6} \cmidrule(lr){7-8}
\textbf{Dataset} & \textbf{Dim} & \textbf{Spearman $\rho$} & \textbf{$p$-value} & $\rho$ & $p$-value & $\rho$ & $p$-value \\
\midrule
Full Gold Std. & 25  & 0.555 & $< 10^{-7}$  & 0.310 & $< 0.003$ & -0.045 & n.s. $p = 0.67$ \\
($N=90$)       & 100 & \textbf{0.593} & $< 10^{-9}$ & 0.177 & n.s., $p = 0.096$ & -0.078 & n.s. $p = 0.47$ \\
               & 300 & 0.567 & $< 10^{-8}$  & 0.170 & n.s. $p = 0.110$ & -0.080 & n.s. $p = 0.45$ \\
\midrule
High Consensus & 25  & 0.716 & $< 10^{-3}$  & 0.371 & n.s. $p = 0.129$ & 0.027 & n.s. $p = 0.91$ \\
($N=18$)       & 100 & 0.800 & $< 10^{-4}$  & 0.208 & n.s. $p = 0.408$ & 0.066 & n.s. $p = 0.80$ \\
               & 300 & \textbf{0.807} & $< 10^{-4}$  & 0.230 & n.s. $p = 0.358$ & 0.066 & n.s. $p = 0.80$ \\
\bottomrule
\end{tabular*}
\label{tab:main_results}
\end{table}

\textbf{Architectural Impact and Sparsity:} The ``dimensional collapse'' observed in Word2Vec is evident as performance degrades by 43\% between $25d$ ($\rho = 0.310$) and $100d$ ($\rho = 0.177$). More importantly, while the $25d$ model shows a weak correlation, the $100d$ representation fails to reach statistical significance ($p > 0.05$), highlighting the inability of sliding-window methods to estimate stable vectors as the parameter-to-data ratio increases in small corpora.
In Yucatec Maya, data sparsity prevents the convergence of high-dimensional manifolds. Conversely, Node2Vec leverages global graph topology through random walks, allowing it to capture ``latent" semantic paths that are invisible to local context windows.

\textbf{The Failure of Sub-word Information:} Despite its effectiveness in morphologically rich languages with large datasets, FastText yielded negative or near-zero correlations ($\rho \approx -0.080$). This suggests that standard character $n$-grams ($3$--$6$ characters) introduce excessive noise in Mayan languages, where morphemic boundaries are highly variable and the total token count is insufficient to learn productive patterns.

\textbf{Consensus and Model Upper Bound:} 

\begin{figure}[t]
    \centering    \includegraphics[width=1.0\textwidth]{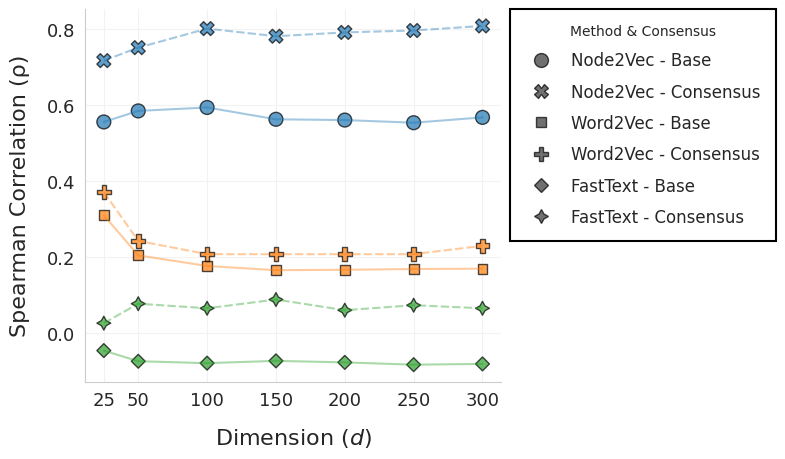}
    \caption{Spearman correlation ($\rho$) across embedding dimensions
($d \in \{25, 50, 100, 150, 200, 250, 300\}$) for Node2Vec (blue),
Word2Vec (orange), and FastText (green). Solid lines show results on the full gold standard ($N = 90$); dashed lines show results on the
high-consensus subset ($N = 18$). Node2Vec yields higher and more stable correlations than the distributional baselines on both sets.}
    \label{fig:semantic_stability}
\end{figure}

The impact of inter-annotator standard deviation is visualized in Figure \ref{fig:semantic_stability}.
Our analysis reveals a significant performance leap when focusing on a subset of 18 high-consensus pairs (those with an inter-annotator standard deviation of $\sigma < 0.5$ or absolute consensus). Within this group, the Spearman correlation increases from $\rho = 0.567$ to $\rho = 0.807$, representing a 42.3\% relative improvement. 
More importantly, this transition corresponds to a substantial increase in the squared Spearman correlation ($\rho^2$), rising from 0.321 to 0.651. This indicates that the proportion of shared rank-order variance between the model and human semantic judgments increases from approximately 32\% to 65\%, suggesting a markedly stronger alignment with the core semantic structure reflected in these instances.
The model effectively captures both strong semantic ties, such as the link between \textit{ch'úupal} (girl) and \textit{winiko'ob} (people) (with an average human rating of $3.8$), and the lack of association in unrelated pairs like \textit{paakat} (look) and \textit{wak'aj} (swollen), which experts unanimously rated as $0.0$.
%
%
%
%

\section{Discussion and Limitations}

The consistent superiority of Node2Vec over distributional baselines suggests that for languages with high morphological dispersion, the global topology of a co-occurrence graph is a more resilient signal than local context density. 
These results are consistent with the findings of \citeauthor{gutierrezvasques2017lowresourcebilinguallexiconextraction}, who reported that Node2Vec embeddings learned from a bilingual graph outperformed Word2Vec for Spanish–Nahuatl bilingual lexicon extraction in a low-resource setting. Although their task is cross-lingual lexicon induction rather than monolingual semantic similarity, both studies suggest that graph-structured training can help compensate for data scarcity in Indigenous languages of Mexico. The present study extends this evidence to Yucatec Maya.
In Yucatec Maya, a single root can generate dozens of inflected forms; while Word2Vec treats these as independent, sparse entities, the random walks in Node2Vec act as a ``semantic bridge'' that aggregates these occurrences into structural paths.

\begin{figure}[t]
\centering
\begin{tikzpicture}[
    node distance=0.5cm,
    matrix_cell/.style={rectangle, draw=gray!30, minimum size=0.55cm, font=\tiny},
    zero_cell/.style={fill=white},
    stable_cell/.style={fill=blue!60}, 
    unstable_cell/.style={fill=blue!15}, 
    word_node/.style={circle, draw=blue!60, fill=blue!10, inner sep=2pt, font=\scriptsize\bfseries, minimum size=1cm},
    highlight_path/.style={draw=orange, line width=1.5pt, -{Stealth}, bend left},
]

\matrix (sparse_matrix) [matrix of nodes, nodes={matrix_cell}, column sep=-\pgflinewidth, row sep=-\pgflinewidth] {
|[zero_cell]|     & |[unstable_cell]| & |[zero_cell]|     & |[zero_cell]|     & |[zero_cell]| \\ 
|[unstable_cell]| & |[zero_cell]|     & |[unstable_cell]| & |[stable_cell]|   & |[unstable_cell]| \\ 
|[zero_cell]|     & |[unstable_cell]| & |[zero_cell]|     & |[zero_cell]|     & |[zero_cell]| \\ 
|[zero_cell]|     & |[stable_cell]|   & |[zero_cell]|     & |[zero_cell]|     & |[stable_cell]| \\ 
|[zero_cell]|     & |[unstable_cell]| & |[zero_cell]|     & |[stable_cell]|   & |[zero_cell]| \\ 
};

\foreach \word [count=\i] in {xa'an, k'áax, naaj, che', si'} {
    \node[left=0.1cm of sparse_matrix-\i-1, font=\tiny\itshape, anchor=east] {\word};
}
\node[below=0.1cm of sparse_matrix, font=\tiny, text=gray] {Distributional Sparsity};

\node[word_node] (n1) at (7, 0.8) {k'áax};
\node[word_node] (n2) at (5, 0) {xa'an};
\node[word_node] (n3) at (9, 0) {naaj};
\node[word_node] (n4) at (6, -1.2) {che'};
\node[word_node] (n5) at (8, -1.2) {si'};

\draw[gray!30, line width=0.4pt] (n2) -- (n1); 
\draw[gray!30, line width=0.4pt] (n3) -- (n1); 
\draw[gray!60, line width=1.2pt] (n4) -- (n1); 
\draw[gray!30, line width=0.4pt] (n5) -- (n1); 
\draw[gray!60, line width=1.2pt] (n4) -- (n5); 

\draw[highlight_path] (n2) to (n1);
\draw[highlight_path] (n1) to (n3);

\node[orange, font=\scriptsize\bfseries, text width=2.5cm, align=center] at (7, -0.2) {Relational Bridge\\(Inferred Similarity)};

\draw[dashed, gray!20] (3.5, -1.8) -- (3.5, 1.8);

\end{tikzpicture}
\caption{Visualizing the convergence gap. Distributional models (left) rely on high-intensity co-occurrence signals (dark blue) to achieve statistical convergence. Low-frequency terms produce weak, unstable signals (light blue), leading to fragmented semantic spaces. Our graph approach (right) leverages the fixed topology to infer relationships through bridges, even when individual connections are weak.}
\label{fig:relational_bridge}
\end{figure}
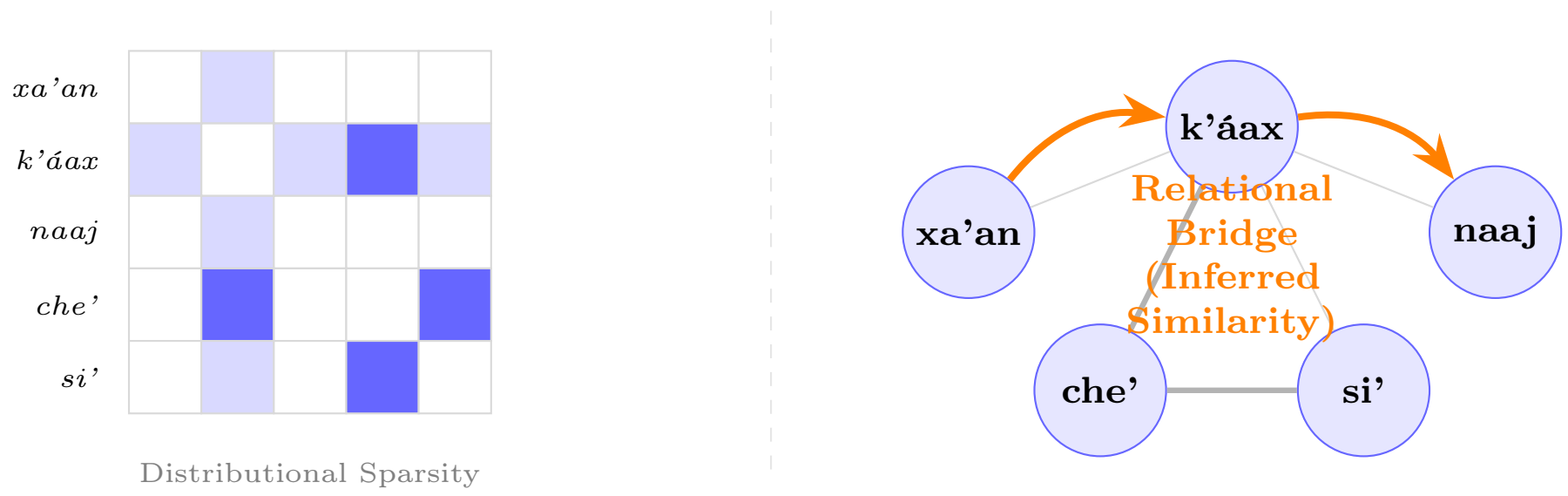

As illustrated in Figure \ref{fig:relational_bridge}, distributional models (left) grapple with the data sparsity problem. In these models, the lack of frequent direct co-occurrences within local windows leads to unstable vector convergence, resulting in a fragmented semantic space for rare terms. Conversely, our relational approach (right) mitigates this by constructing a global co-occurrence graph. This topological structure enables the recovery of semantic proximity between low-frequency terms (e.g., xa’an and naaj) through multi-hop paths and common neighbors such as k’áax, bypassing the need for high-frequency statistical reinforcement.
The observed ``Consensus Gain'', a 42\% improvement in correlation when filtering for unambiguous pairs, indicates that once human uncertainty is removed, the graph's topological structure aligns closely with expert linguistic intuition.

Our findings further gain significance when contrasted with existing benchmarks for low-resource American languages. \citet{gongora-etal-2022-use} report a maximum Spearman correlation of 0.569 for Guaraní using a translated version of the MC-30 dataset. It is important to note that although Guaraní is an Indigenous language, it is also one of the official languages of Paraguay. This institutional status has facilitated the availability of a comparatively larger amount of textual data than is typically found for digitally low-resource languages. Despite ongoing challenges such as orthographic standardization and variation, this relatively broader corpus availability constitutes a crucial factor for training reliable word embedding models.
Furthermore, \citet{molina2025generating} highlight the extreme sensitivity of Mayan embeddings to hyperparameter tuning when using distributional models. Their analysis reveals that while Swadesh lists maintain stable performance ($\rho \approx 0.57-0.58$), the MC-30 dataset is highly volatile, with performance increasing by roughly 75\% (from 0.16 to 0.28) only through intensive window-size optimization. 
Crucially, they identify negative sampling ($nsamp$) as the most influential factor (impact = 0.236), while noting that increasing dimensionality offers minimal benefits ($\rho \approx 0.20$ for MC-30). 
It is important to emphasize that both Swadesh and MC-30 are originally English-based lists; thus, their translated counterparts may not fully capture the language-specific semantic nuances of Yucatec Maya.

In contrast to the iterative statistical tuning required by distributional baselines, our graph-based approach involves a higher computational cost during the construction and traversal of the global co-occurrence network, which precluded an equivalent exhaustive hyperparameter grid search. 

However, our results suggest that the global topological signal is comparatively robust. Without an exhaustive hyperparameter search, Node2Vec reaches $\rho = 0.567$--$0.593$ on the full 90-pair gold standard. This is comparable to the median correlation reported for Swadesh-based evaluation of Yucatec Maya embeddings ($\rho \approx 0.57$) \citep{molina2025generating} and to the best MC-30 result for Guaraní ($\rho \approx 0.569$) \citep{gongora-etal-2022-use}, although below the best tuned model in \citet{molina2025generating} ($\rho \approx 0.733$). On the 18-pair high-consensus subset, the correlation rises to $\rho \approx 0.807$. These comparisons should nonetheless be interpreted with caution: the benchmarks differ in language, size, construction (translated word lists versus native expert ratings), and the origin of the reference scores, and the high-consensus subset is small.


\subsection{Morphological Noise in FastText} 
The failure of FastText ($\rho \approx -0.08$) is a critical finding. In high-resource settings, sub-word $n$-grams capture morphology; however, in our 717k-word corpus, these fragments appear to introduce noise. Without enough data to filter out coincidental character overlaps, the model assigns similarity to words that share affixes but have unrelated meanings, contradicting native speaker perception.

\subsection{Limitations}
Despite its success, the study faces the following limitations:

\begin{itemize}
    \item \textbf{Representational Balance:} While the corpus includes everyday conversational data via the T'aantsil source, the distribution remains predominantly journalistic. This imbalance may still bias the embeddings toward formal structures over spontaneous oral syntax.
    \item \textbf{Orthographic and Dialectal Variation:} Although we implemented rigorous normalization, Yucatec Maya exhibits high regional variation. Our corpus, while multi-genre, may not fully capture the specificities of all dialectal variants across the Yucatán peninsula.
    \item \textbf{Computational Cost:} The Node2Vec training required significant resources, preventing an exhaustive search of the $p$ (return) and $q$ (in-out) hyperparameters. While our results are significant, further optimization of these parameters could yield even higher alignments.
\end{itemize}

\section{Conclusion and Future Work}

Returning to the initial question, what do Mayan embeddings capture? Our results demonstrate that while traditional distributional models are limited by surface frequencies, a graph-based approach successfully captures the discourse relationships hidden in the long-tail of Yucatec Maya. As our lead linguist noted, the model recovers associations based on discourse context, rather than strictly grammatical assignments. We have introduced the first human-rated gold standard for evaluating semantic embeddings in this language, demonstrating that graph topologies are particularly well-suited to overcoming the data scarcity inherent in morphologically complex, low-resource environments. By publishing our Gold Standard of 90 pairs, we provide a critical resource for the digital revitalization of the Mayan language family.

Future work could focus on digitizing oral traditions to balance the current journalistic dominance with spontaneous conversational data, capturing a broader range of discourse relationships.
We plan to develop architectures that explicitly combine graph topology with morphological segmentation of subwords. This aims to leverage the advantages of character-level models like FastText while maintaining the structural stability provided by global graph embeddings.
Beyond intrinsic similarity tasks, we intend to evaluate these representations in applied contexts such as Mayan-Spanish machine translation and cross-lingual information retrieval.
Our research marks a significant step towards linguistic equity in NLP. By prioritizing structural depth and native expert intuition over raw data volume, we demonstrate that indigenous languages can benefit from modern computational tools.

\section*{Funding}
This work was supported by the Secretariat of Science, Humanities, Technology, and Innovation (SECIHTI) of Mexico, under projects INMX-4974 and CBF-2025-G-368.

\section*{Conflict of Interest}
The authors declare no financial, professional, or personal conflicts of interest that could influence the results or the publication of this work.

\section*{Acknowledgments}
The authors wish to express their deep gratitude to Lorena Pool Balam, MA in Mayan Psycholinguistics, for her leadership as principal evaluator and expert curator of the Gold Standard. We also sincerely thank our fellow professors, linguists, and native Mayan speakers Jesus Edilberto Sunza Pech, Rosa Couoh Pool, Didier Chan Quijano, and Irma Pomol Cahum. Their linguistic expertise and profound cultural knowledge were fundamental in executing the blind evaluation protocol and ensuring the quality of this study.

\bibliography{sn-bibliography}

\end{document}